\documentclass[sigconf]{acmart}
\AtBeginDocument{%
  }

\usepackage{multirow} 
\usepackage{threeparttable}
\usepackage{balance}
\copyrightyear{2025}
\acmYear{2025}
\setcopyright{acmlicensed}
\acmConference[MM '25] {Proceedings of the 33rd ACM International Conference on Multimedia}{October 27--31, 2025}{Dublin, Ireland.}
\acmBooktitle{Proceedings of the 33rd ACM International Conference on Multimedia (MM '25), October 27--31, 2025, Dublin, Ireland}
\acmISBN{979-8-4007-2035-2/2025/10}
\acmDOI{10.1145/3746027.3755835}

\begin{document}

\title{TeleAntiFraud-28k: An Audio-Text Slow-Thinking Dataset for Telecom Fraud Detection}


\author{Zhiming Ma}
\authornote{Both authors contributed equally to this research.}
\authornote{Corresponding author}
\email{mazhiming312@outlook.com}
\orcid{0009-0004-5955-7978}
\affiliation{%
  \institution{China Mobile Internet Company Ltd.}
  \city{Guangzhou}
  \country{China}
}

\author{Peidong Wang}
\authornotemark[1]
\authornotemark[2]
\email{pdongwang@163.com}
\orcid{0009-0008-5015-1065}
\affiliation{%
  \institution{Northeastern University}
  \city{Shenyang}
  \country{China}
}

\author{Minhua Huang}\orcid{0009-0001-3325-6032}
\author{Jinpeng Wang}\orcid{0009-0001-3501-2760}
\affiliation{%
  \institution{China Mobile Internet Company Ltd.}
  \city{Guangzhou}
  \country{China}
}

\author{Kai Wu}\orcid{0009-0002-3749-6268}
\author{Xiangzhao Lv}\orcid{0009-0008-6989-3747}
\affiliation{%
  \institution{China Mobile Internet Company Ltd.}
  \city{Guangzhou}
  \country{China}
}

\author{Yachun Pang}\orcid{0009-0007-3939-7251}
\author{Yin Yang}\orcid{0009-0004-9655-7432}
\affiliation{%
  \institution{China Mobile Internet Company Ltd.}
  \city{Guangzhou}
  \country{China}
}

\author{Wenjie Tang}\orcid{0009-0004-0625-659X}
\author{Yuchen Kang}\orcid{0009-0007-6390-9050}
\affiliation{%
  \institution{China Mobile Internet Company Ltd.}
  \city{Guangzhou}
  \country{China}
}

\renewcommand{\shortauthors}{Zhiming Ma et al.}

\begin{abstract}
The detection of telecom fraud faces significant challenges due to the lack of high-quality multimodal training data that integrates audio signals with reasoning-oriented textual analysis. To address this gap, we present TeleAntiFraud-28k, the first open-source audio-text slow-thinking dataset specifically designed for automated telecom fraud analysis. Our dataset is constructed through three strategies: (1) Privacy-preserved text-truth sample generation using automatically speech recognition-transcribed call recordings (with anonymized original audio), ensuring real-world consistency through text-to-speech model regeneration; (2) Semantic enhancement via large language model based self-instruction sampling on authentic ASR outputs to expand scenario coverage; (3) Multi-agent adversarial synthesis, which simulates emerging fraud tactics through predefined communication scenarios and fraud typologies, enriches the conversation samples. The generated dataset contains 28,511 rigorously processed audio-text pairs with a total audio duration of more than 307 hours, complete with detailed annotations for fraud reasoning. The dataset is divided into three tasks: scenario classification, fraud detection, fraud type classification. Furthermore, we construct TeleAntiFraud-Bench, a standardized evaluation benchmark comprising proportionally sampled instances from TeleAntiFraud-28k, to facilitate systematic testing of model performance, reasoning capabilities, and thought processes on telecom fraud detection tasks. We also contribute a supervised fine-tuning model based on Qwen2-Audio, trained on the TeleAntiFraud-28k training set, while open-sourcing the data processing framework to enable community-driven dataset expansion. This work establishes a foundational framework for multimodal anti-fraud research while addressing critical challenges in data privacy and scenario diversity. The code of this paper is publicly available at \url{https://github.com/JimmyMa99/TeleAntiFraud}.
\end{abstract}




\maketitle

\begin{figure}[h]
  \centering
\includegraphics[width=0.76\linewidth]{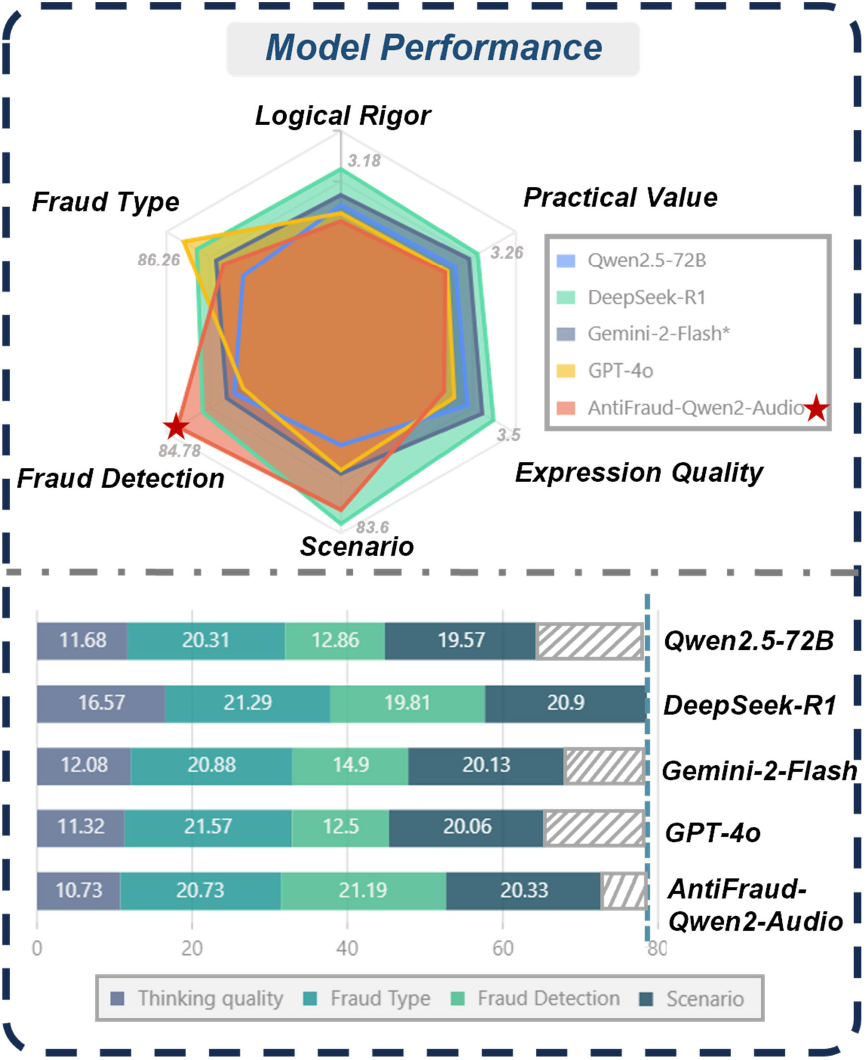}
  \caption{Models' Performance on TeleAntiFraud-Bench}
  \label{fig:case_study}
   
\end{figure}

\begin{figure*}[th]
    \centering
    \includegraphics[width=0.55\linewidth]{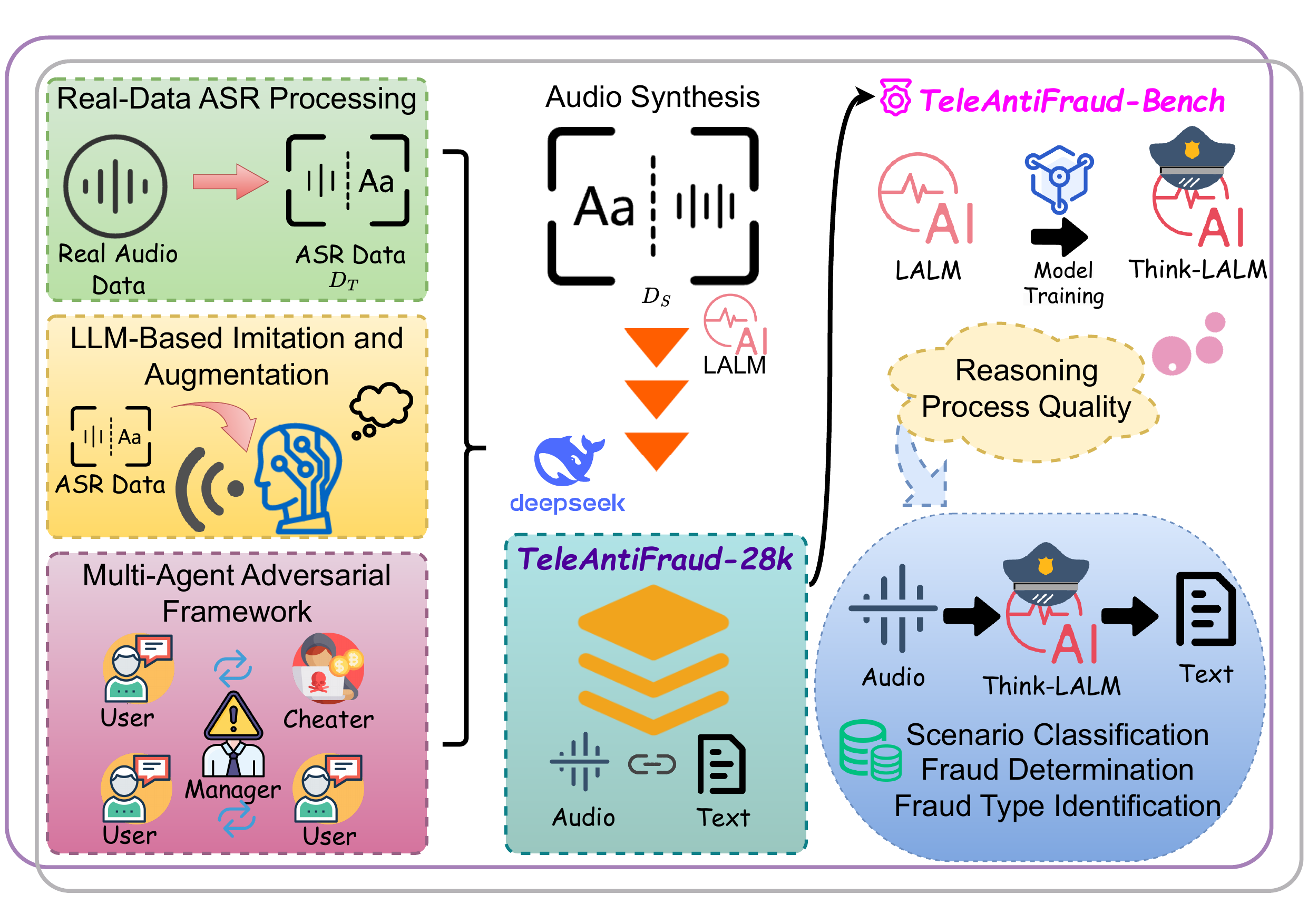}
    
    \caption{\textbf{An overview of TeleAntiFraud-28k.} Our system addresses telecom fraud detection challenges by creating TeleAntiFraud-28k through Real-Data ASR Processing, LLM-Based Imitation and Augmentation, and multi-agent adversarial synthesis. We develop TeleAntiFraud-Bench for evaluation and provide a supervised fine-tuning model with open-sourced data processing.}
    \label{fig:speech-generation}
    
\end{figure*}

\section{Introduction}
As telecom fraud techniques grow increasingly sophisticated, they pose escalating threats to social security and economic stability. Global economic losses fraud-related have reached \$1.02 trillion, representing 1.05\% of the global GDP, a significant increase over 2020-2021 figures, with over a quarter of respondents reporting encounters with fraud \cite{GASA2024}.  Developing effective detection methods has therefore become urgent. Traditional approaches relying on manual verification and rule-based pattern matching offer limited accuracy and adaptability against rapidly evolving fraudulent strategies.

Recent advancements in large language models (LLMs), particularly their slow-thinking reasoning capabilities \cite{min2024imitate}, offer promising solutions for combating telecom fraud. However, a significant modality gap exists between voice calls (the primary source of fraud data) and the text data that LLMs process, limiting their direct application. Current industry methods typically employ automatic speech recognition (ASR) to convert audio into text before applying carefully designed prompts for LLM-based fraud detection \cite{jiang2024detecting}. Although demonstrating partial effectiveness, this methodology relies extensively on precise prompt engineering, exhibiting significant performance variability across diverse models and implementation contexts. Furthermore, the ASR conversion process often results in information loss \cite{li2021icassp}, potentially omitting crucial fraud indicators contained in vocal features such as tone and pauses that frequently signal fraudulent intent.

Slow-thinking reasoning enhances both accuracy and interpretability of LLMs in judgment tasks. Concurrently, emerging large audio language models (LALMs) capable of directly processing audio signals offer potential solutions to the modality gap in anti-fraud applications. However, the absence of slow-thinking audio datasets specifically designed for telecom fraud constrains the application and performance improvement of LALMs in fraud detection.

To address this gap, this study proposes TeleAntiFraud-28k, an audio-text slow-thinking dataset for telecom fraud detection. The dataset is constructed through three methodologies: 1) Transcribing anonymized call recordings using ASR technology to generate privacy-protected text samples, then employing text to speech models to regenerate samples that align with real-world language expressions while preserving content authenticity and safeguarding the privacy of both conversation parties; 2) Implementing self-instructed sampling strategies with LLMs to semantically enhance content, further expanded through TTS augmentation techniques; 3) Designing a multi-agent adversarial framework that simulates emerging fraud tactics through predefined communication scenarios and fraud typologies, enabling the expansion and generation of novel fraudulent scripts.TeleAntiFraud-28k provides detailed slow-thinking annotations covering communication scenario classification, fraud determination, and fraud type identification, designed to enhance model explainability and accuracy through training while improving the model's capabilities in understanding conversation contexts, detecting fraudulent activities, and categorizing fraud types respectively.

We systematically extract representative samples to construct TeleAntiFraud-Bench, an evaluation benchmark preserving the original dataset's scenario distribution and fraud type proportions, ensuring reliable assessment outcomes. This benchmark provides researchers with a unified evaluation platform for comparative analysis across telecommunication scenario classification, fraud detection, fraud type classification tasks, and assessment of model reasoning processes, enabling comprehensive evaluation of both outcomes and thought patterns.

Experiments on state-of-the-art Large Audio Language Models (LALMs) demonstrate that current LALMs without fine-tuning perform inadequately for telecom anti-fraud tasks. After fine-tuning Qwen2-Audio on the TeleAntiFraud-28k training set, we observed a 27.5-point score increase on TeleAntiFraud-Bench, demonstrating the dataset's effectiveness and practical value in developing audio-based anti-fraud models.

The contributions of this research can be summarized as follows:

\begin{enumerate}
    \item Proposing the first multi-task slow-thinking audio-language dataset TeleAntiFraud-28k for telecom fraud prevention, encompassing three tasks: communication scenario classification, fraud determination, and fraud type analysis;
    \item Designing a new data generation pipeline that maximizes coverage of diverse fraud scenarios through real-call ASR processing, LLM-based simulation, and multi-agent adversarial generation;
    \item Establishing the TeleAntiFraud-Bench evaluation benchmark to provide standardized testing standards for telecom fraud detection models, while designing a series of evaluations for anti-telecom fraud slow-thinking capabilities;
    \item Conducting comprehensive evaluations on multiple leading-edge LALMs using TeleAntiFraud-Bench, validating the dataset's training effectiveness and establishing performance baselines for future audio-based anti-fraud research.
\end{enumerate}

    
    
    

This work bridges critical research gaps in multimodal fraud detection and provides valuable resources for advancing intelligent anti-fraud systems. The dataset and benchmark are publicly available to facilitate community-wide research efforts in combating evolving telecom fraud.

\section{Related work}
\subsection{LLM-Based Telecom Fraud Detection}
Large language models have been applied to telecom fraud detection, but existing research primarily focuses on text analysis. Singh et al.~ \cite{singh2025advanced} proposed a RAG-based system achieving 97.98\% accuracy but relies on text analysis alone. Shen et al.~ \cite{shen2025warned} developed a real-time framework for fraudulent intent detection that remains text-centric. Shen et al.~ \cite{shen2024combating} identified LLM challenges including data bias and hallucinations in fraud detection. Chang~ \cite{10825256} revealed LLM vulnerabilities to adversarial scams, though limited to textual.

Current anti-fraud systems rely on predefined rules~ \cite{ahmed2021semantic} or text-only analysis without audio integration~ \cite{mbaziira2016text}. Trained on limited datasets, these systems struggle with evolving fraud tactics, highlighting the need for a large-scale, multimodal telecom fraud dataset.

\subsection{Multimodal audio-Text Models}

Advances in deep learning have enabled significant progress in multimodal models, yet these innovations remain underutilized in telecom fraud detection \cite{bello2024artificial}.

Recent large multimodal models demonstrate tremendous potential in telecom fraud prevention. GLM-4-Voice~ \cite{zeng2024glm} supports real-time bilingual voice interaction with adjustable audio characteristics including emotion, tone, and dialect. Qwen2-Audio~ \cite{chu2024qwen2} processes diverse audio inputs for text responses. Among proprietary systems, GPT-4o~ \cite{achiam2023gpt} handles arbitrary combinations of text, audio, image, and video inputs to generate multimodal outputs.

Despite possessing certain capabilities, existing multimodal models lack domain-specific optimization for telecom fraud detection, as they are primarily designed for general conversation or content comprehension. Due to privacy concerns associated with call audio, these models have limited access to call-related training data during their development process. Additionally, challenges such as difficulties in data acquisition, stringent privacy requirements, and the diverse nature of fraud scenarios require specialized approaches that current general-purpose models are not equipped to handle.

While LLM-based fraud detection systems and multimodal audio-text models have advanced independently, their integration for combating telecom fraud remains inadequate. The TeleAntiFraud-28k dataset addresses these limitations by providing large-scale, slow-thinking-annotated audio-text pairs, establishing infrastructure for multimodal anti-fraud research.

\section{Method}

This chapter details the TeleAntiFraud-28k construction methodology and the design of the evaluation framework, TeleAntiFraud-Bench. As depicted in Figure \ref{fig:speech-generation}, our approach encompasses three components: voice data generation, text data annotation based on slow-thinking, and the establishment of an evaluation benchmark.

\begin{figure}[h]
  \centering
\includegraphics[width=0.68\linewidth]{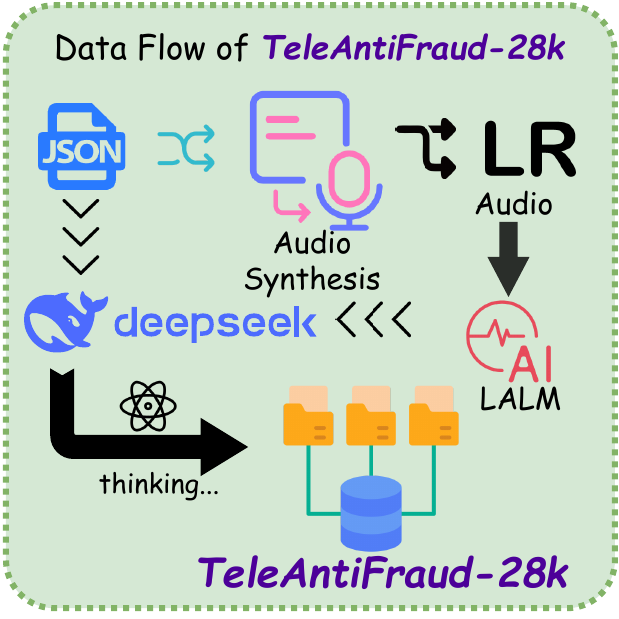}

  \caption{Data Flow of Audio Data Generation}
  \label{fig:dataflow}
   
\end{figure}

Our methodology involves a three-stage process. First, we generate a high-quality, diverse telephone-call dataset covering both telecom fraud and normal scenarios using three novel methods. Next, in a "slow-thinking" annotation phase, an audio-understanding model's analysis and ASR outputs are used to prompt the DeepSeek-R1 large reasoning model, which in turn generates detailed annotations that include its explicit reasoning process. Finally, we establish a unified benchmark to evaluate various models on the telecom anti-fraud task.

\subsection{Audio Data Generation Pipeline}
To construct the high-quality, diverse TeleAntiFraud-28k dataset, we developed a robust voice data generation pipeline. Our approach, illustrated in Figure~\ref{fig:dataflow}, comprises two key stages: dialogue text generation and voice synthesis.

For dialogue text generation, we employed three complementary methods: real data ASR processing, large language model imitation generation, and a multi-agent adversarial framework. These approaches collectively ensured authenticity, diversity, and comprehensive scenario coverage. In the subsequent voice synthesis stage, we converted the generated dialogue texts into dual-channel voice recordings using TTS technology, creating separate audio tracks for callers and callees that simulate realistic telephone conversations.

\noindent\textbf{Real-Data ASR Processing.}
Our first dialogue acquisition approach leverages real telephone recordings through ASR processing. We collected actual telecom fraud recordings ($D^F_T$) and normal calls ($D^N_T$), then converted these into separate channel texts using ASR technology. This method preserves authentic speech patterns and conversational dynamics while anonymizing sensitive information. The resulting transcripts were subsequently transformed into synthetic voice data using TTS models, producing realistic fraudulent ($D^F_{S1}$) and normal ($D^N_{S1}$) recordings. This approach ensures generated content closely resembles genuine fraud scenarios while addressing privacy concerns inherent in utilizing original recordings directly.

\noindent\textbf{LLM-Based Imitation and Augmentation.}
Our second dialogue generation method utilizes the self-instruct paradigm \cite{wang2022self} to augment ASR-processed call transcripts. We designed prompt templates incorporating few-shot exemplars from our initial datasets, $D^F_{S1}$ and $D^N_{S1}$, to guide the LLM in generating dialogues that are both diverse and pattern-consistent. To ground the generation in realism, prompts were enriched with key linguistic features, interaction patterns, and fraudulent tactics extracted from real calls. For fraudulent content, we preserved core deception strategies while systematically varying circumstantial details. This process yielded two substantial, high-quality datasets of synthetic fraudulent ($D^F_{S2}$) and normal ($D^N_{S2}$) calls. The self-instruct approach proved effective for controllably scaling our data volume and diversity, thereby enhancing model generalization to better simulate real-world scenarios.

\noindent\textbf{Multi-Agent Adversarial Framework.}
To augment conversational text data, we introduce a multi-agent adversarial framework designed to simulate emerging fraud tactics and expand data diversity. We have identified that current telecom fraud datasets, while authentic, exhibit a limited range of conversational patterns and scenarios, hindering their effectiveness against novel fraud techniques. Our proposed framework addresses this by simulating fraudulent activities in varied business contexts, thereby enriching the data distribution to cover a wider array of fraud strategies.

\begin{figure}[h]
  \centering
\includegraphics[width=0.72\linewidth]{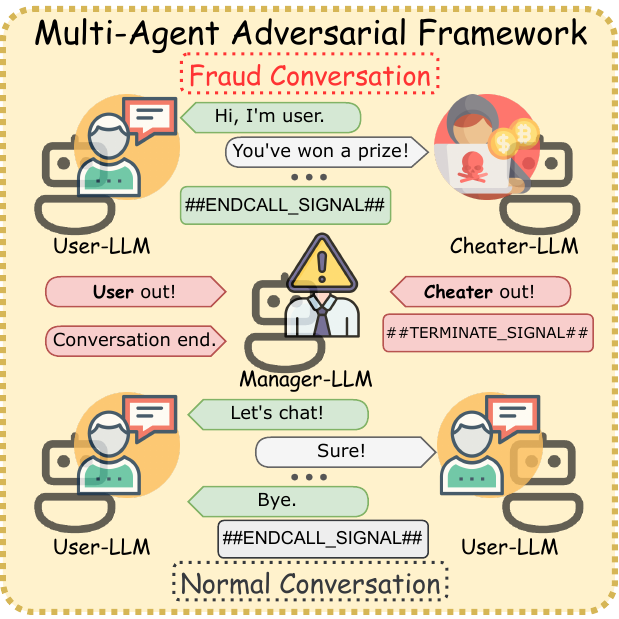}
    
\caption{Structure of Multi-Agent Adversarial Framework}
\label{fig:MAAF}
   
\end{figure}

As illustrated in Figure \ref{fig:MAAF}, the framework involves three agents: a caller (cheater), a callee (potential victim), and a Manager. The caller is given a specific fraud type, and the callee has a defined profile to ensure realistic interactions. The conversation is turn-based, with the Manager monitoring for adherence to the preset scenario and natural flow.

To expand data diversity, we designed numerous scenario-fraud type combinations based on seven normal and eight real-world fraud scenarios. This approach systematically guides the data generation process, effectively broadening the data distribution and filling existing gaps.

A special signal mechanism facilitates natural conversation termination. The caller or callee can send \textbf{\#\#ENDCALL\_SIGNAL\#\#} to hang up, while the Manager can use \textbf{\#\#TERMINATE\_SIGNAL\#\#} to end the dialogue, preventing loops or topic deviation.

Through the multi-agent adversarial framework, we generated a new set of normal call data $D^N_{S3}$ and fraudulent call data $D^F_{S3}$.  The data generated by the multi-agent framework significantly expanded the distribution space of the original data, particularly in new fraud types and complex scenarios, providing more comprehensive training data for the model.

\noindent\textbf{Audio Synthesis.}
After generating dialogue texts, we employed ChatTTS \cite{ChatTTS} to convert them into dual-channel audio. This advanced TTS model excels in conversational scenarios with natural, expressive speech, multi-speaker capabilities, and prosodic control.

To enhance realism, we implemented three strategies:

\noindent\textbf{1. Diverse Audio Characteristics}: We randomized voice parameters (timbre, rate, pitch) and configured distinct characteristics for different speakers.

\noindent\textbf{2. Natural Speech Features}: We preserved pauses, stress, and emotional variations, enhancing deception-specific expressions (urgency, authority, false affinity) for fraudulent scenarios.

\noindent\textbf{3. Precise Temporal Control}: We managed the timing between voices to ensure turn-taking and appropriate response delays.

Our dataset comprises normal calls \(D^N = D^N_{S1} \cup D^N_{S2} \cup D^N_{S3}\) and fraudulent calls \(D^F = D^F_{S1} \cup D^F_{S2} \cup D^F_{S3}\). Original recordings were transcribed, anonymized, and resynthesized as \(D^N_{S1}\) and \(D^F_{S1}\). Our public release includes these anonymized segments only in the test set, alongside LLM-generated and multi-agent-derived data \(D^N_{S2} \cup D^N_{S3}\) and \(D^F_{S2} \cup D^F_{S3}\).

\subsection{Slow Thinking-Based Text Annotation}
The TeleAntiFraud-28k dataset's distinctive feature is its "Slow Thinking" mechanism that emulates anti-fraud experts' analytical processes. We implemented a two-stage workflow for generating high-quality annotations:

\textbf{1. Audio Analysis Phase}: Generated voice data is processed through a professional audio understanding model to extract voice features and information, including emotional variations, tonal characteristics, pause patterns, and other audio-level indicators.

\textbf{2. Expert Reasoning Phase}: Audio analysis results and ASR text are combined as prompts for the DeepSeek-R1, which acts as a voice analysis expert with anti-fraud professional knowledge. The model employs \texttt{"<think></think>"} markers to document its complete reasoning chain from clue identification to judgment, and \texttt{"<answer></answer>"} to denote final conclusions.

This design captures anti-fraud experts' systematic framework, encompassing speech pattern recognition, fraud technique identification, and risk assessment. Our designed prompt template presents conversation content and audio feature analysis, requiring the model to output JSON-formatted information including reasoning grounds (reason), confidence level (confidence), and a boolean fraud determination (is\_fraud).

The annotation process involves three sequential analytical steps:

\textbf{1. Call Scene Classification}: The model analyzes the conversation's basic context and theme, categorizing it into one of seven predefined scenarios: "Dining Service", "Customer Consultation", "Appointment Service", "Transportation Inquiry", "Routine Shopping", "Ride-Hailing Service", or "Food Delivery Service".

\textbf{2. Fraud Determination}: Building on scene classification, the model evaluates potential fraudulent behavior by analyzing speech characteristics, request reasonableness, information disclosure patterns, and other professional indicators, providing a clear judgment with supporting evidence.

\textbf{3. Fraud Type Identification}: For identified fraudulent calls, the model categorizes them into seven main types: "Investment Fraud," "Phishing Fraud", "Identity Theft", "Lottery Fraud", "Banking Fraud", "Extortion Fraud", or "Customer Service Fraud".

Each analytical step yields JSON output with judgment parameters and confidence levels. This progressive analysis mirrors professional workflows, building context to enhance accuracy and reliability, while the documented reasoning provides a rich dataset for future research.

\subsection{Construction of Evaluation Benchmark}
To systematically evaluate model performance in telecom anti-fraud tasks, we established the TeleAntiFraud-Bench evaluation benchmark. This benchmark comprises samples randomly extracted from the TeleAntiFraud-28k dataset in proportion, incorporating genuine normal data ($D^N_{S1}$) and genuine fraud data ($D^F_{S1}$) into the test set. This construction methodology preserves the original dataset's scenario and fraud type distributions, ensuring evaluation results maintain high representativeness and reliability.

\noindent\textbf{Design of the Evaluation Process.}
TeleAntiFraud-Bench employs a hybrid evaluation mechanism combining rule-based extraction with LLM analysis to ensure comprehensive and accurate assessment. We designed structured prompts to guide evaluation LLMs while simultaneously utilizing regular expressions for precise information extraction. This approach comprises four key steps:

\textbf{1. Regular Expression-Based Key Information Extraction}: We extract critical results from model outputs—including scenario classification, fraud judgment, and fraud type—using regular expressions. This approach provides precise foundational data for subsequent evaluation, ensuring accuracy and consistency independent of LLM analysis.

\textbf{2. LLM Analysis of the Thought Process}: The LLM evaluation analyzes the model's reasoning process (content within \texttt{"<think></think>"} markers) for completeness, logical coherence, and effective use of evidence, using the language understanding and analytical capabilities of the LLM.

\textbf{3. LLM Verification of Final Judgment Consistency}: The evaluation LLM verifies alignment between the model's final judgment and standard answers, utilizing its strengths in logical judgment and comparative analysis to ensure reliable evaluation results.

\begin{figure}[h]
 
  \centering
\includegraphics[width=0.95\linewidth]{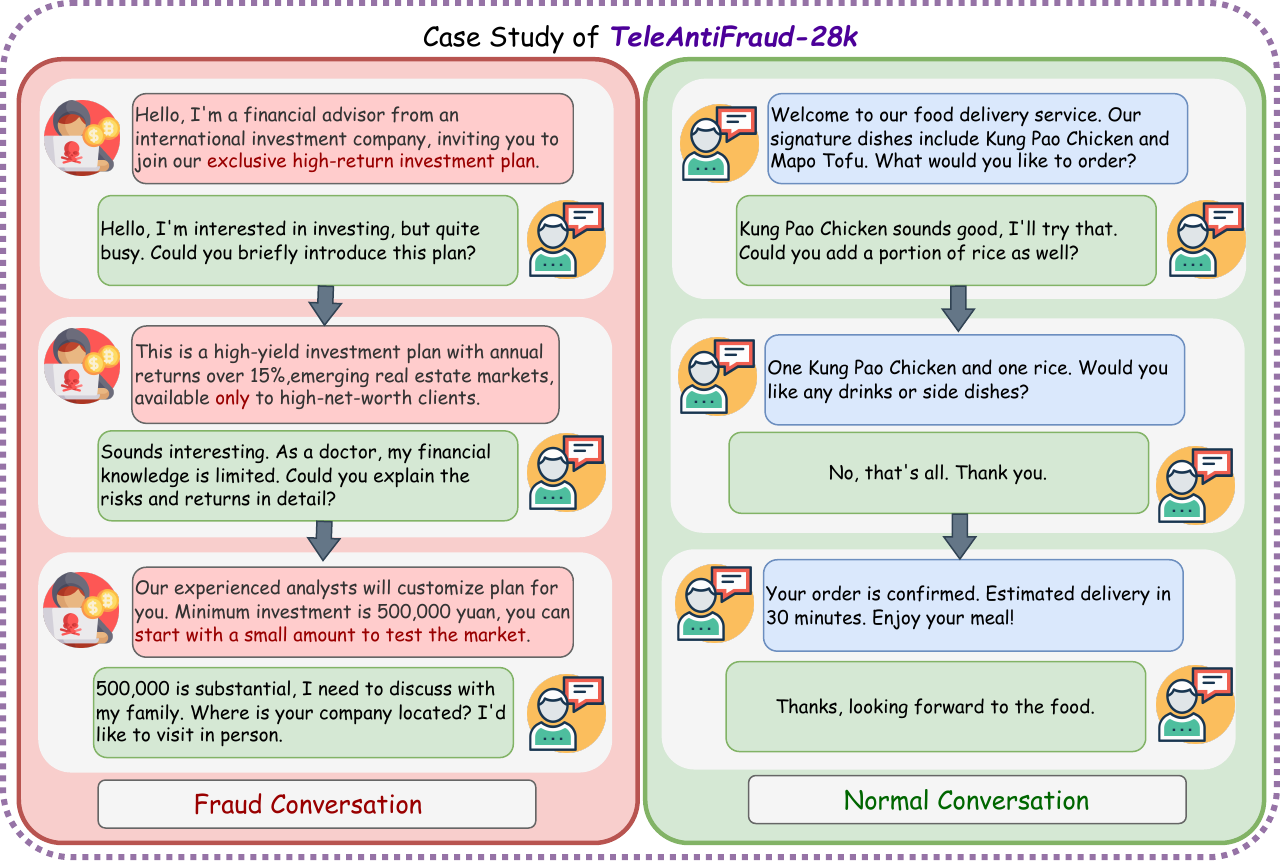}
  
  \caption{Case Study of TeleAntiFraud-28k}
  \label{fig:case_study}
   
\end{figure}

\textbf{4. Comprehensive Evaluation Score and Report Generation}: We integrate regular expression-extracted information with LLM analysis to generate comprehensive evaluation scores and detailed analytical reports that thoroughly present model performance across multiple dimensions.

This combined methodology improves evaluation robustness by integrating pattern-matching precision from regular expressions with the contextual understanding of LLMs, providing more thorough and accurate assessment of model performance.

\noindent\textbf{Inference Evaluation Process.}
To comprehensively assess reasoning quality, we developed an evaluation framework based on three dimensions: logical consistency, practicality, and clarity. This framework employs structured prompting \cite{wang2024langgpt} to guide an LLM in performing scoring tasks using model reasoning ($R_m$), model answer ($A_m$), reference answer ($A_r$), and reference reasoning ($R_r$). The scoring process combines expert rules and logical constraints through a probabilistic mechanism.

Logical consistency evaluates reasoning chain completeness, assumption reasonableness, and conclusion derivation rigor. When $A_m \neq A_r$, scoring probabilities significantly decrease, with further reductions for each logical leap or missing evidence. When $R_m \neq R_r$ but $A_m = A_r$, probabilities are moderately reduced to account for reasoning divergence while acknowledging the possibility of multiple valid logical approaches. Practicality assesses problem essence understanding, solution effectiveness, and demand coverage completeness. If $A_m \neq A_r$, probabilities for problem identification and solution effectiveness default to zero. Even with matching answers, reasoning differences trigger verification of solution pathway validity against established domain principles. Clarity measures key point presentation, language conciseness, and information organization efficiency, with probability decay triggered by unclear or redundant expressions. When reasoning differs, additional scrutiny of expression efficiency and organization occurs, with probability adjustments proportional to the degree of structural and presentational divergence from reference reasoning.

We allocate scores using probabilistic distributions that quantify the likelihood of different values. For a scoring point with possible values $S \in \{0, 1, 2\}$, we assign probability $P(S)$ and calculate the expected value using $E = \sum_{S} S \cdot P(S)$. The total score combines expected values from logical consistency ($E_L$), practicality ($E_U$), and clarity ($E_C$):
\[ E_{\text{total}} = E_L + E_U + E_C \]

This evaluation system integrates quantitative metrics with probabilistic calculations for objective and comprehensive reasoning quality assessment.

\noindent\textbf{Evaluation Metric System.}
TeleAntiFraud-Bench employs a balanced scoring mechanism with four equally weighted dimensions (25\% each). We use weighted F1 scores as our quantitative metrics, calculated as:
$$F1_{task} = \sum_{i = 1}^{n} w_{task_i} \times \frac{2 \times P_{task_i} \times R_{task_i}}{P_{task_i} + R_{task_i}}$$
Where $w_{task_i}=\frac{n_{i}}{\sum_{j = 1}^{n}n_{j}}$ represents the proportion of samples in class $i$, $P_{task_i}$ and $R_{task_i}$ denote precision and recall for class $i$, and $n$ is the total number of classes.

Our evaluation framework comprises four key metrics:

\textbf{1. Scene Classification F1 Score}: Measures accuracy in identifying seven call scenarios, using precision $P_{scene_i}$ and recall $R_{scene_i}$ across scene categories.

\textbf{2. Fraud Judgment F1 Score}: Evaluates accuracy in determining call legitimacy, using precision $P_{fraud_i}$ and recall $R_{fraud_i}$ for fraud detection.

\textbf{3. Fraud Type Identification F1 Score}: Assesses accuracy in categorizing seven fraud types, using precision $P_{type_i}$ and recall $R_{type_i}$ across fraud categories.

\textbf{4. Quality of Thought Process}: Evaluated by an LLM examining completeness, evidence utilization, professional knowledge application, and logical coherence.

The comprehensive scoring system combines multiple evaluation dimensions with equal weighting. The total score is calculated as:$Score_{total} = 0.25 \times F1_{scene} + 0.25 \times F1_{fraud} + 0.25 \times F1_{type} + 0.25 \times E_{total}$

This balanced approach integrates scene identification, fraud detection, type classification, and process evaluation metrics to provide a complete assessment of system performance.

\section{Evaluation}
\subsection{Experimental Setup}

\noindent\textbf{Dataset Statistics.}The constructed TeleAntiFraud-28k dataset comprises 28,511 utterance samples, which are divided into training and test sets. The training set contains 21,490 samples, constituting 75.38\% of the total dataset, while the test set contains 7,021 samples, constituting 24.62\%. This 3:1 partition ratio ensures sufficient samples for benchmark evaluation while adhering to the conventional training-test split protocol widely adopted in machine learning methodology. Detailed examples from the dataset are illustrated in Figure \ref{fig:case_study}. Table~\ref{tab:call-distribution} presents the  statistical information of the dataset.

\begin{table}[h]
\caption{Distribution of Scam and Normal Calls in the Dataset}
 
\small
\centering
\begin{tabular}{lccc}
\hline
\textbf{Type} & \textbf{Total} & \textbf{Fraud Calls} & \textbf{Normal Calls} \\
\hline
Train& 21,490 & 9,950 (46.3\%) & 11,540 (53.7\%) \\
Test& 7,021 & 3,697 (52.66\%) & 3,324 (47.34\%) \\
Total & 28,511 & 13,647 (47.86\%) & 14,864 (52.13\%) \\
\hline
\end{tabular}
\label{tab:call-distribution}
\end{table}

\begin{figure*}[th]
    \centering    \includegraphics[width=0.8\linewidth]{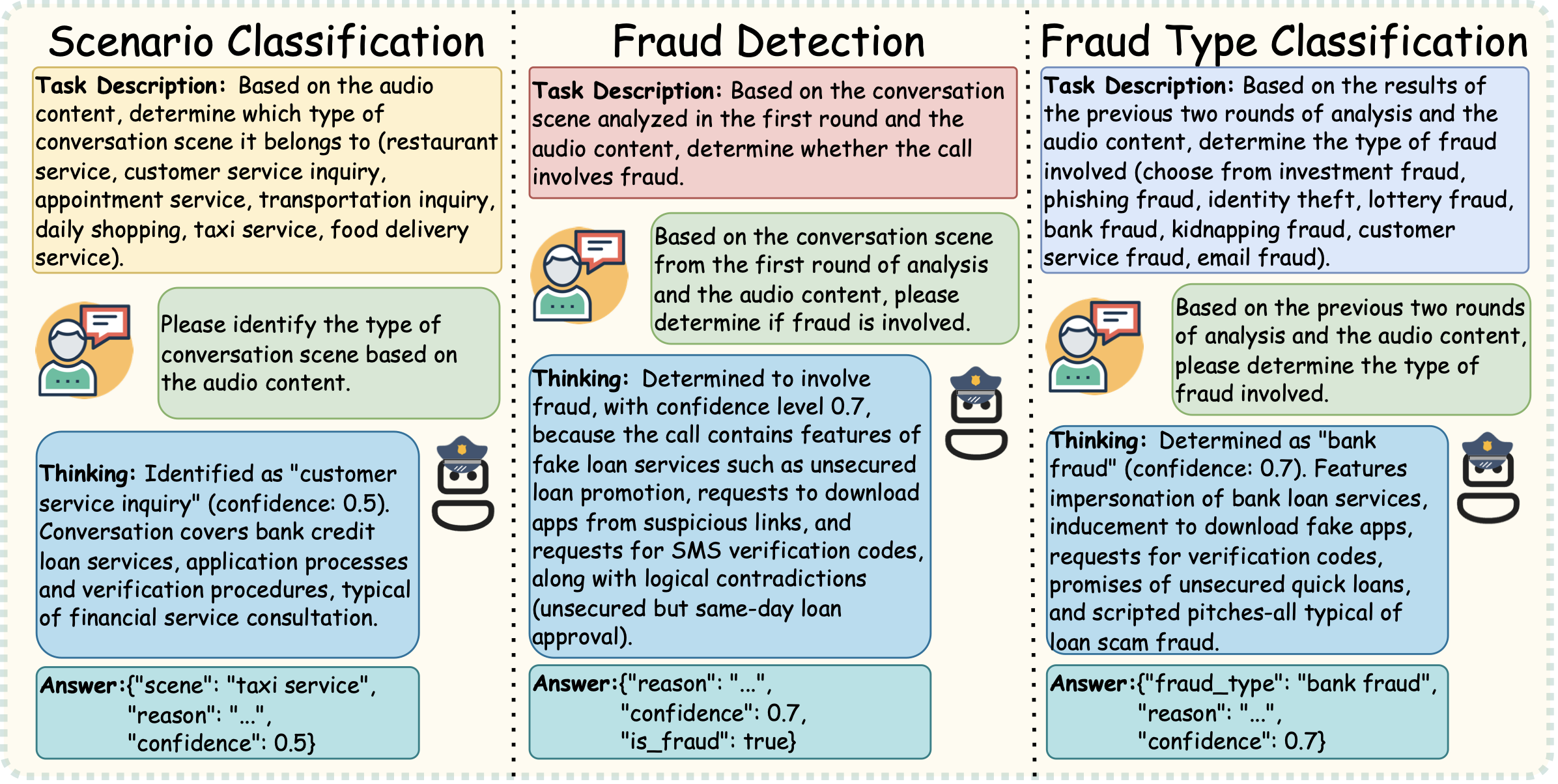}
    
    \caption{Case Study of Multi-Task Learning}
    \label{fig:case-task}
\end{figure*}

The TeleAntiFraud-28k maintains relative balance between fraudulent and normal calls, with fraud calls constituting 46.3\% (9,950 samples) of the training set and 52.66\% (3,697 samples) of the test set. Fraud calls represent 47.86\% (13,647 samples) of the entire data set, ensuring a balanced evaluation of detection capabilities.

\begin{table}[htbp]
\centering
\caption{Distribution of scenario types in train and test sets}
 
\small
\label{tab:scenario-distribution}
\begin{tabular}{lrr}
\toprule
\textbf{Scenario Type} & \textbf{Training Set} & \textbf{Test Set} \\
\midrule
Customer Consultation & 6,421 & 4,632 \\
Appointment Services & 1,714 & 867 \\
Routine Shopping & 924 & 340 \\
Dining Services & 581 & 154 \\
Food Delivery Services & 494 & 448 \\
Ride-Hailing Services & 353 & 489 \\
Transportation Inquiries & 223 & 91 \\
\midrule
\textbf{Total} & 10,710 & 7,021 \\
\bottomrule
\end{tabular}
 
\end{table}

The TeleAntiFraud-28k dataset was built to be diverse and representative of various conversational scenarios. After rigorous selection and annotation, the training set features multiple interaction types (Table \ref{tab:scenario-distribution}), with customer consultation being the largest category (6,421 samples), followed by appointment services (1,714 samples). This diversity is mirrored in the test set to validate robust model generalization. This approach provides a solid foundation for practical applications, such as the Multi-Task Learning case studies illustrated in Figure \ref{fig:case-task}.

In terms of fraud types, the dataset covers seven major categories: phishing, kidnapping, lottery, customer service, banking, investment, and identity theft. As shown in Table\ref{tab:fraud-distribution}, the sample distribution reflects real-world occurrence patterns, with variations that align with practical fraud detection challenges.

\begin{table}[htbp]
\centering
\caption{Distribution of fraud types in train and test sets}

\small
\label{tab:fraud-distribution}
\begin{tabular}{lrr}
\toprule
\textbf{Fraud Type} & \textbf{Training Set} & \textbf{Test Set} \\
\midrule
Customer Service Fraud & 2,022 & 725 \\
Banking Fraud & 1,626 & 2,408 \\
Investment Fraud & 785 & 216 \\
Phishing Fraud & 443 & 123 \\
Lottery Fraud & 418 & 99 \\
Kidnapping Fraud & 324 & 91 \\
Identity Theft & 105 & 35 \\
\midrule
\textbf{Total} & 5,723 & 3,697 \\
\bottomrule
\end{tabular}
 
\end{table}

\noindent\textbf{Model Selection for Evaluation.}
In telecom fraud detection, ASR combined with LLMs is the predominant approach. This study evaluated 10 representative models on the TeleAntiFraud-28k dataset, using SenseVoice as the standardized ASR. For ASR+LLM evaluation, we selected high-performance open-source models: DeepSeek-V3, DeepSeek-R1, Doubao-1.5-Pro, InternLM2.5-20B-Chat, GLM-4-9B-Chat, and Qwen2.5-72B-Instruct. Multimodal evaluation included commercial models (GPT-4o, Gemini-2.0-Flash) and open-source options (GLM-4-Voice, Step-1o-audio, Qwen2-Audio). These selections span diverse scales and types, ensuring comprehensive evaluation.

\subsection{Experimental Results}

\begin{table*}[htbp]
\centering
\caption{Comprehensive Evaluation of Different Models on Quality Metrics and Fraud Detection Tasks}
 
\small
\label{tab:combined-model-evaluation}
\begin{tabular*}{\textwidth}{@{\extracolsep{\fill}}lcccccccc@{}}
\toprule
\multirow{2}{*}{\textbf{Model}} & \multicolumn{4}{c}{\textbf{Quality Evaluation}} & \multicolumn{3}{c}{\textbf{Fraud Detection Performance}} & \multirow{2}{*}{\textbf{Final}} \\
\cmidrule(lr){2-5} \cmidrule(lr){6-8}
& \textbf{Logical} & \textbf{Practical} & \textbf{Expression} & \textbf{Total} & \textbf{Scenario} & \textbf{Fraud} & \textbf{Fraud Type} & \\
& \textbf{Rigor} & \textbf{Value} & \textbf{Quality} & \textbf{(15)} & \textbf{(\%)} & \textbf{(\%)} & \textbf{(\%)} & \\
\midrule
\multicolumn{9}{c}{\textbf{ASR+LLM}} \\
GLM4-9B-Chat \cite{glm2024chatglm} & 1.61 & 1.43 & 2.20 & 5.25 & 75.10 & 46.91 & 82.22 & 59.81 \\
InternLM2.5-20B \cite{cai2024internlm2} & 1.99 & 1.93 & 2.43 & 6.37 & 78.34 & 36.67 & 85.42 & 60.72 \\
Qwen2.5-72B \cite{qwen2.5} & 2.21 & 2.16 & 2.70 & 7.01 & 78.31 & 51.44 & 81.24 & 64.43 \\
DeepSeek-V3 \cite{deepseekai2024deepseekv3technicalreport} & 2.32 & 2.34 & 2.85 & 7.51 & 88.53 & 14.62 & 66.71 & 54.98 \\
Doubao-1.5-Pro \cite{DoubaoTeamDoubao15Pro2025} & 1.94 & 1.75 & 2.60 & 6.31 & 71.14 & 36.11 & 82.25 & 57.89 \\
\midrule
\multicolumn{9}{c}{\textbf{ASR+Reasoning LLM}} \\
DeepSeek-R1 \cite{deepseekai2025deepseekr1incentivizingreasoningcapability} & 3.18 & 3.26 & 3.50 & 9.94 & 83.60 & 79.25 & 85.16 & 78.57 \\
\midrule
\multicolumn{9}{c}{\textbf{LALM}} \\
Qwen2-Audio-7B-Instruct & 1.51 & 1.42 & 1.96 & 4.91 & 70.22 & 58.51 & 20.47 & 45.48 \\
GLM4-9B-Voice* & 0.89 & 0.64 & 0.65 & 1.89 & - & 26.83 & 38.33 & - \\
Gemini-2-Flash* \cite{GoogleDeepMindGemini2024} & 2.25 & 2.29 & 2.72 & 7.25 & 80.51 & 59.61 & 83.53 & 68.00 \\
GPT-4o* \cite{achiam2023gpt} & 2.12 & 2.10 & 2.56 & 6.79 & 80.25 & 50.00 & 86.26 & 65.44 \\
Step-1o-Audio* \cite{huang2025stepaudiounifiedunderstandinggeneration} & 1.64 & 1.62 & 2.01 & 5.26 & 76.35 & 40.65 & 79.71 & 57.94 \\
\midrule
\multicolumn{9}{c}{\textbf{Fine-tuned Anti-Fraud LALM}} \\
AntiFraud-Qwen2-Audio & 2.06 & 2.07 & 2.31 & 6.44 & 81.31 & 84.78 & 82.91 & 72.98 \\
\bottomrule
\end{tabular*}
\begin{tablenotes}
\small
\item Note: Scenario = Scene Classification F1, Fraud = Fraud Detection F1, Classification F1, AVG F1= Average Score, Total(15) = Quality of Thought Process Score, *: The test samples are 1,200 pieces selected through sampling.
\end{tablenotes}
 
\end{table*}

\noindent\textbf{Scenario Classification Performance.}
In scenario classification, DeepSeek-V3 achieved the highest F1 score (88.53\%), demonstrating ASR+LLM advantages, followed by DeepSeek-R1 (83\%). The fine-tuned AntiFraud-Qwen2-Audio (81.31\%) exhibited stable performance, confirming that anti-fraud-optimized language models effectively enhance task performance. Multimodal models performed reliably, with GPT-4o (80.25\%) and Gemini-2.0-Flash (80.51\%) providing comprehensive information retrieval, although falling short of the top ASR + LLM models. InternLM2.5-20B-Chat (78. 34\%) and GLM-4-9B-Chat (75. 1\%) highlighted the limitations of language models in scenario classification.

\noindent\textbf{Fraud Detection Performance.}
Fraud detection tasks revealed more pronounced performance differences. DeepSeek-R1 excelled with an F1 score of 79. 25\%, showcasing the strong recognition capabilities of the ASR+slow-thinking LLM approach. The fine-tuned AntiFraud-Qwen2-Audio model performed best (84.78\%), demonstrating that specially optimized speech-language models enhance fraud detection reliability. Multimodal models showed varied performance: GPT-4o (50\%) and Step-1o-audio (40.65\%) underperformed compared to ASR+LLM approaches, indicating challenges in integrating audio and language information. Gemini-2.0-Flash (59.61\%) demonstrated balanced performance, while InternLM2.5-20B-Chat (36.67\%) and GLM-4-Voice (26.83\%) revealed significant limitations in practical fraud detection.

\noindent\textbf{Fraud Type Classification Performance.}
In fraud type classification, models performed similarly. DeepSeek-R1 led with an F1 score of 85.16\%, followed by InternLM2.5-20B-Chat (85.42\%) and Doubao-1.5-Pro (82.25\%). AntiFraud-Qwen2-Audio achieved 82.91\%, while without fine-tuning it scored only 58.51\%, demonstrating fine-tuning's effectiveness. The multimodal GPT-4o performed exceptionally (86.26\%), showing advantages in multimodal information integration, while Gemini-2.0-Flash achieved 74.55\%.

\noindent\textbf{Quality Assessment of the Thinking Process.}
We conducted a comprehensive evaluation of model reasoning quality across three dimensions—logical rigor, practical value, and expressive quality—using DeepSeek-R1 as a systematic judge (Table \ref{tab:combined-model-evaluation}). As predicted by scaling law theory, a model's reasoning ability is strongly correlated with its parameter count, which explains the superior performance of specialized reasoning models like DeepSeek-R1 and DouBao-1.5-Pro.

Notably, our AntiFraud-Qwen2-Audio model outperformed both similarly-sized models (GLM4-9B-Chat, GLM4-9B-Voice) and significantly larger ones (InternLM2.5-20B, Step-Audio-Chat 130B), achieving a reasoning quality comparable to GPT-4o. This result clearly demonstrates the efficacy of our task-specific optimization. In contrast, the untrained Qwen2-Audio-7B-Instruct exhibited substantially lower reasoning quality, underscoring the critical importance of our slow-thinking training set for cultivating the model's advanced analytical capabilities.

\noindent\textbf{Comprehensive Evaluation Model.}
In the telecom fraud detection task, our comprehensive evaluation revealed AntiFraud-Qwen2-Audio as the leading performer with an 83\% average F1 score, validating our proposed data synthesis and fine-tuning strategies. Analysis of reasoning abilities showed this model excelled in thinking process quality within the DeepSeek-R1 evaluation framework, significantly outperforming competitors in logical rigor (2.06), practical value (2.07), and expressive quality (2.31), achieving a total score of 6.44. This performance surpassed similarly sized models like GLM4-9B-Chat and approached GPT-4o's capabilities. The untrained Qwen2-Audio-7B-Instruct scored only 4.91 in thinking quality evaluation, while the slow-thinking trained AntiFraud-Qwen2-Audio demonstrated significant improvement, validating our training strategy. The multimodal GPT-4o* achieved a 72.17\% average F1 score, highlighting both potential and challenges of multimodal approaches, while InternLM2.5-20B-Chat and GLM-4-9B-Chat showed relatively weaker performance (66.81\% and 68.08\% respectively). These results underscore the importance of targeted data synthesis, and slow-thinking training for enhancing model performance in telecom fraud detection.

\subsection{Ablation Studies}

\noindent\textbf{Audio's Influence on Model Judgment Capability.}
We analyzed audio and text features' impact on model performance across two dimensions. As shown in Table \ref{tab:model-metrics-comparison}, models trained solely on ASR text achieved an average F1 score of 73.58\%, substantially outperforming the base model (49. 73\%), indicating that text features effectively capture conversational semantic information. The Without Think and Think multimodal models achieved average F1 scores of 73.93\% and 83.00\% respectively, demonstrating that multimodal fusion significantly enhances fraud detection capabilities. Audio feature integration supplements text with vocal nuances, enhancing the model's fraud scenario comprehension.

\begin{table}[htbp]
\centering
\small
\caption{F1 Comparison of Different Qwen2-Audio Variants}

\label{tab:model-metrics-comparison}
\begin{tabular}{lcccc}
\toprule
\textbf{Variant} & \textbf{Scene} & \textbf{Fraud} & \textbf{Type} & \textbf{Avg F1} \\
\midrule
Base & 70.22 & 58.51 & 20.47 & 49.73 \\
ASR-text & 71.55 & 71.27 & 77.93 & 73.58 \\
NO Think & 72.08 & 69.32 & 80.39 & 73.93 \\
Think & 81.31 & 84.78 & 82.91 & 83.00 \\
\bottomrule
\end{tabular}
\begin{tablenotes}
\small
\item Note: Scenario = Scene Classification F1, Fraud = Fraud Detection F1, Classification F1, AVG F1= Average Score. Base = untrained model, ASR-text = model with only speech ASR text without audio, NO Think = model without slow thinking process, Think = model with slow thinking.
\end{tablenotes}
\vspace{-1em}
\end{table}

\noindent\textbf{Influence of Slow Thinking on the Performance of the Model.}
The slow-thinking mechanism represents a key innovation in this study. Comparative analysis between models trained with Without Think and Think datasets demonstrates significant performance enhancements when incorporating slow-thinking strategies. The F1 score improved from 72.08\% to 80.91\% for scenario classification, from 69.32\% to 84.78\% for fraud involvement detection, and from 80.39\% to 82.91\% for fraud type classification. This substantial improvement (average F1 score increase from 73.93\% to 83.00\%) validates the value of slow thinking in telecom fraud detection—the multi-step reasoning mechanism enables deeper analysis of call scenarios, captures more potential fraud features, and reduces hasty judgments. Additionally, the slow-thinking approach enhances system explainability and credibility, aligning more closely with human anti-fraud expert reasoning.

\section{Conclusion}
This study addresses telecom fraud detection challenges by creating the TeleAntiFraud-28k dataset through our three-phase strategy: Real-Data ASR Processing, LLM-Based Imitation, and Multi-Agent Adversarial Framework. This first slow-thinking dataset for telecom fraud comprises 28,511 audio-text pairs with fraud reasoning annotations. We established TeleAntiFraud-Bench to evaluate models and trained AntiFraud-Qwen2-Audio, demonstrating improved fraud detection through multimodal fusion and structured thinking. Our approach provides foundations for future systems with real-world anti-fraud applications.

\section*{Acknowledgements}
Thanks to all co-authors for their hard work. We would like to express our sincere gratitude to China Mobile Internet Company, NEU Data Mining, ModelScope Community, and SmartFlowAI Community for their valuable support and assistance throughout the course of this work.
Special thanks go to Qingyun Pan, Wenxing Hu and Jintao Huang for their insightful guidance and helpful contributions.
The work is supported by the Natural Science Foundation of China (62272092, 62172086).

\bibliographystyle{ACM-Reference-Format}
\balance
\bibliography{TeleAntiFraud}

\begin{thebibliography}{10}

\bibitem{GASA2024}
Jorij Abraham, Sam Rogers, Luka Koning, Clement Njoki, and James Greening.
\newblock \emph{Global State of Scams Report 2024}, March 2025.
\newblock \url{https://www.gasa.org/_files/ugd/7bdaac_9060be8317424edd9964072cf279a0a4.pdf}.
\newblock Accessed: March 25, 2025.

\bibitem{singh2025advanced}
Gurjot Singh, Prabhjot Singh, and Maninder Singh.
\newblock Advanced real-time fraud detection using rag-based llms.
\newblock \emph{arXiv preprint arXiv:2501.15290}, 2025.

\bibitem{shen2025warned}
Zitong Shen, Sineng Yan, Youqian Zhang, Xiapu Luo, Grace Ngai, and Eugene Yujun Fu.
\newblock ``It warned me just at the right moment'': Exploring llm-based real-time detection of phone scams.
\newblock \emph{arXiv preprint arXiv:2502.03964}, 2025.

\bibitem{shen2024combating}
Zitong Shen, Kangzhong Wang, Youqian Zhang, Grace Ngai, and Eugene Y. Fu.
\newblock Combating phone scams with llm-based detection: Where do we stand?
\newblock \emph{arXiv preprint arXiv:2409.11643}, 2024.

\bibitem{10825256}
Chen-Wei Chang, Shailik Sarkar, Shutonu Mitra, Qi Zhang, Hossein Salemi, Hemant Purohit, Fengxiu Zhang, Michin Hong, Jin-Hee Cho, Chang-Tien Lu, et~al.
\newblock Exposing llm vulnerabilities: Adversarial scam detection and performance.
\newblock In \emph{2024 IEEE International Conference on Big Data (BigData)}, pages 3568--3571, 2024.

\bibitem{zeng2024glm}
Aohan Zeng, Zhengxiao Du, Mingdao Liu, Kedong Wang, Shengmin Jiang, Lei Zhao, Yuxiao Dong, and Jie Tang.
\newblock Glm-4-voice: Towards intelligent and human-like end-to-end spoken chatbot.
\newblock \emph{arXiv preprint arXiv:2412.02612}, 2024.

\bibitem{chu2024qwen2}
Yunfei Chu, Jin Xu, Qian Yang, Haojie Wei, Xipin Wei, Zhifang Guo, Yichong Leng, Yuanjun Lv, Jinzheng He, Junyang Lin, et~al.
\newblock Qwen2-audio technical report.
\newblock \emph{arXiv preprint arXiv:2407.10759}, 2024.

\bibitem{achiam2023gpt}
Josh Achiam, Steven Adler, Sandhini Agarwal, Lama Ahmad, Ilge Akkaya, Florencia Leoni Aleman, Diogo Almeida, Janko Altenschmidt, Sam Altman, Shyamal Anadkat, et~al.
\newblock Gpt-4 technical report.
\newblock \emph{arXiv preprint arXiv:2303.08774}, 2023.

\bibitem{an2024funaudiollm}
Keyu An, Qian Chen, Chong Deng, Zhihao Du, Changfeng Gao, Zhifu Gao, Yue Gu, Ting He, Hangrui Hu, Kai Hu, et~al.
\newblock Funaudiollm: Voice understanding and generation foundation models for natural interaction between humans and llms.
\newblock \emph{arXiv preprint arXiv:2407.04051}, 2024.

\bibitem{chen2024bge}
Jianlv Chen, Shitao Xiao, Peitian Zhang, Kun Luo, Defu Lian, and Zheng Liu.
\newblock Bge m3-embedding: Multi-lingual, multi-functionality, multi-granularity text embeddings through self-knowledge distillation.
\newblock \emph{arXiv preprint arXiv:2402.03216}, 2024.

\bibitem{deepseekai2024deepseekv3technicalreport}
DeepSeek-AI.
\newblock \emph{DeepSeek-V3 Technical Report}, 2024.
\newblock \url{https://arxiv.org/abs/2412.19437}.

\bibitem{deepseekai2025deepseekr1incentivizingreasoningcapability}
DeepSeek-AI.
\newblock \emph{DeepSeek-R1: Incentivizing Reasoning Capability in LLMs via Reinforcement Learning}, 2025.
\newblock \url{https://arxiv.org/abs/2501.12948}.

\bibitem{cai2024internlm2}
Zheng Cai, Maosong Cao, Haojiong Chen, Kai Chen, Keyu Chen, Xin Chen, Xun Chen, Zehui Chen, Zhi Chen, Pei Chu, et~al.
\newblock \emph{InternLM2 Technical Report}, 2024.
\newblock \url{https://arxiv.org/abs/2403.17297}.

\bibitem{glm2024chatglm}
Team GLM, Aohan Zeng, Bin Xu, Bowen Wang, Chenhui Zhang, Da Yin, Diego Rojas, Guanyu Feng, Hanlin Zhao, Hao Yu, et~al.
\newblock \emph{ChatGLM: A Family of Large Language Models from GLM-130B to GLM-4 All Tools}, 2024.
\newblock \url{https://arxiv.org/abs/2406.12793}.

\bibitem{qwen2.5}
An Yang, Baosong Yang, Beichen Zhang, Binyuan Hui, Bo Zheng, Bowen Yu, Chengyuan Li, Dayiheng Liu, Fei Huang, Haoran Wei, et~al.
\newblock \emph{Qwen2.5 Technical Report}, 2024.
\newblock \url{https://arxiv.org/abs/2412.15115}.

\bibitem{team2023gemini}
Team, Gemini, Rohan Anil, Sebastian Borgeaud, Jean-Baptiste Alayrac, Jiahui Yu, Radu Soricut, Johan Schalkwyk, Andrew M. Dai, Anja Hauth, Katie Millican, et~al.
\newblock \emph{Gemini: A Family of Highly Capable Multimodal Models}, 2023.
\newblock \url{https://arxiv.org/abs/2312.11805}.

\bibitem{huang2025stepaudiounifiedunderstandinggeneration}
Ailin Huang, Boyong Wu, Bruce Wang, Chao Yan, Chen Hu, Chengli Feng, Fei Tian, Feiyu Shen, Jingbei Li, Mingrui Chen, et~al.
\newblock \emph{Step-Audio: Unified Understanding and Generation in Intelligent Speech Interaction}, 2025.
\newblock \url{https://arxiv.org/abs/2502.11946}.

\bibitem{wang2024langgpt}
Ming Wang, Yuanzhong Liu, Xiaoming Zhang, Songlian Li, Yijie Huang, Chi Zhang, Daling Wang, Shi Feng, and Jigang Li.
\newblock \emph{LangGPT: Rethinking Structured Reusable Prompt Design Framework for LLMs from the Programming Language}, 2024.
\newblock \url{https://arxiv.org/abs/2402.16929}.

\bibitem{jiang2024detecting}
Liming Jiang.
\newblock Detecting scams using large language models.
\newblock \emph{arXiv preprint arXiv:2402.03147}, 2024.

\bibitem{min2024imitate}
Yingqian Min, Zhipeng Chen, Jinhao Jiang, Jie Chen, Jia Deng, Yiwen Hu, Yiru Tang, Jiapeng Wang, Xiaoxue Cheng, Huatong Song, et~al.
\newblock Imitate, explore, and self-improve: A reproduction report on slow-thinking reasoning systems.
\newblock \emph{arXiv preprint arXiv:2412.09413}, 2024.

\bibitem{li2021icassp}
Andong Li, Wenzhe Liu, Xiaoxue Luo, Chengshi Zheng, and Xiaodong Li.
\newblock Icassp 2021 deep noise suppression challenge: Decoupling magnitude and phase optimization with a two-stage deep network.
\newblock In \emph{ICASSP 2021 - IEEE International Conference on Acoustics, Speech and Signal Processing (ICASSP)}, pages 6628--6632, 2021.

\bibitem{mbaziira2016text}
A. Mbaziira and J. Jones.
\newblock A text-based deception detection model for cybercrime.
\newblock In \emph{Int. Conf. Technol. Manag}, pages 1--8, 2016.

\bibitem{ahmed2021semantic}
Mansoor Ahmed, Kainat Ansar, Cal B. Muckley, Abid Khan, Adeel Anjum, and Muhammad Talha.
\newblock A semantic rule based digital fraud detection.
\newblock \emph{PeerJ Computer Science}, 7:e649, 2021.

\bibitem{bello2024artificial}
Oluwabusayo Adijat Bello and Komolafe Olufemi.
\newblock Artificial intelligence in fraud prevention: Exploring techniques and applications challenges and opportunities.
\newblock \emph{Computer Science \& IT Research Journal}, 5(6):1505--1520, 2024.

\bibitem{DoubaoTeamDoubao15Pro2025}
DOUBAO TEAM.
\newblock \emph{Doubao-1.5-pro}, January 2025.
\newblock \url{https://team.doubao.com/en/special/doubao_1_5_pro}.
\newblock Accessed: April 8, 2025.

\bibitem{GoogleDeepMindGemini2024}
Google DeepMind.
\newblock \emph{Introducing Gemini 2.0: Our New AI Model for the Agentic Era}, December 2024.
\newblock \url{https://blog.google/technology/google-deepmind/google-gemini-ai-update-december-2024/}.
\newblock Accessed: April 8, 2025.

\bibitem{wang2022self}
Yizhong Wang, Yeganeh Kordi, Swaroop Mishra, Alisa Liu, Noah A. Smith, Daniel Khashabi, and Hannaneh Hajishirzi.
\newblock Self-instruct: Aligning language models with self-generated instructions.
\newblock \emph{arXiv preprint arXiv:2212.10560}, 2022.

\end{thebibliography}

\end{document}